\documentclass{article}
\usepackage{spconf, amsmath, graphicx}
\usepackage{amssymb}
\usepackage{subfigure}
\usepackage{listings}
\usepackage{float}
\usepackage{xcolor}
\usepackage{url}
\usepackage{cite}
\usepackage[ruled,vlined]{algorithm2e}
\usepackage[capitalise]{cleveref}
\usepackage{dblfloatfix}
\usepackage{ragged2e}
\usepackage{setspace}
\graphicspath{{Figures/}}

\SetKwInput{KwInput}{Input}
\SetKwInput{KwOutput}{Output}   

\newcommand{\norm}[1]{\left\lVert #1 \right\rVert}
\newcommand\blfootnote[1]{
	\begingroup
	\renewcommand\thefootnote{ }\footnote{#1}
	\addtocounter{footnote}{-1}
	\endgroup
}

\title{GRAPH-BASED FUSION FOR CHANGE DETECTION IN MULTI-SPECTRAL IMAGES}
\name{\begin{tabular}{c}David Alejandro Jimenez-Sierra$^{\dagger}$, Hern\'an Dar\'io Ben\'itez-Restrepo$^{\dagger}$, Hern\'an Dar\'io Vargas-Cardona$^{\dagger}$ \\ Jocelyn Chanussot$^{\star}$\end{tabular}} 

\address{$^{\dagger}$ Pontificia Universidad Javeriana Cali\\
	Departamento de Electr\'onica y Ciencias de la Computaci\'on\\
	\{davidjimenez, hbenitez, hernan.vargas\}@javerianacali.edu.co \\
	$^{\star}$ Grenoble Images Parole Signals Automatique Laboratory (GIPSA-Lab)\\
Grenoble Institute of Technology\\
jocelyn.chanussot@gipsa-lab.grenoble-inp.fr}

\begin{document}

%\ninept
%
\maketitle

\begin{figure*}[!b]
	\centering
	\includegraphics[scale=0.34]{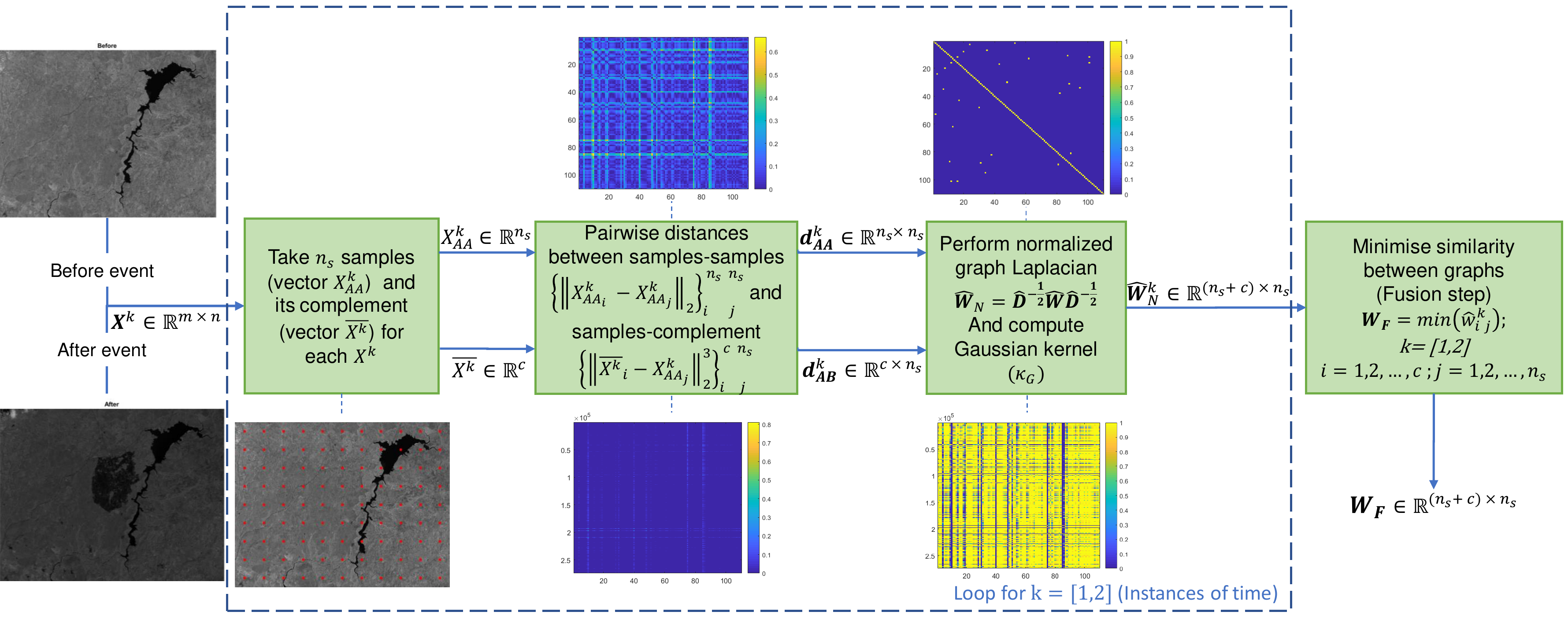}
	\caption{Graph-based fusion. Where, $k$ is the time of the event $1$ (pre) and $2$ (post), $X^{k}$ is an image that represents a event, $X_{AA}^{k}$ represents the samples from $X^{k}$, $\overline{X}^{k}$ is the complement, $\mathbf{d_{AA}}^{k}$ is the pairwise distance between the samples in $X_{AA}^{k}$, $\mathbf{d_{AB}}^{k}$ is the pairwise distance between $X_{AA}^{k}$ and $\overline{X}^{k}$, $\mathbf{\widehat{D}} = Diag(d_{1},d_{2},\dots,d_{n_{s}})\text{ with }d_{i} = \sum_{j}^{n_{s}}\hat{w}_{ij}^{k}$ is the approximated degree matrix , and $\mathbf{\widehat{W}}_{N}^{k}$ is the normalized laplacian calculated by using the Nystr\"om approximation.}
	\label{fig:MMG-chart}
\end{figure*}	

\begin{abstract}
	
In this paper we address the problem of change detection in multi-spectral images by  proposing a data-driven framework of graph-based data fusion. The main steps of the proposed approach are: (i) The generation of a multi-temporal pixel based graph, by the fusion of intra-graphs of each temporal data; (ii) the use of Nystr\"om extension to obtain the eigenvalues and eigenvectors of the fused graph, and the selection of the final change map. We validated our approach in two real cases of remote sensing according to both qualitative and quantitative analyses. The results confirm the potential  of the proposed  graph-based change detection algorithm outperforming state-of-the-art methods.

%The results show that the fusion of temporal data based on graphs detects changes in remote sensing images with high accuracy in percentage with respect to missed alarms($\mathbf{4.8504}$), false alarms ($\mathbf{0.3120}$), precision ($\mathbf{x0.9029}$), recall ( $\mathbf{0.9515}$), Cohen's Kappa ($\mathbf{0.9242}$) and overall error ($\mathbf{0.4463}$), outperforming in most of these metrics the state of the art methods for change detection.

\end{abstract}

\vspace{-0.2cm}

\section{Introduction}
\label{sec:intro}
Change detection (CD) refers to the task of analyzing two or more images acquired over the same area at different times (i.e., multitemporal images) in order to detect zones in which the land-cover type changed between the acquisitions \cite{dalla2015challenges}. CD permits to quantify the magnitude of natural disasters (i.e floodings) and changes generated by  human activity. This analysis provides fundamental data for environmental protection, sustainable development, and maintenance of ecological balance \cite{lahat2015multimodal}. One of the most known sources of data for change detection are the Multi-spectral (MS) images that contain information from both spatial and spectral domain (i.e. Landsat series of satellites). Giving two or more co-registered images, pixel based approaches carry out change detection by probabilistic thresholding and machine learning methods  \cite{yavariabdi2017change,song2018change}. Even though threshold methods are efficient and useful, they are sensitive to MS image noise and require a high accuracy in the estimation of the difference image probabilistic distribution. These issues make threshold methods prone to artifacts in the final change map \cite{kittler1986minimum,zanetti2015rayleigh,zanetti2017theoretical,mian2018new,touati2019multimodal}. Machine learning approaches are divided into two categories: classification and clustering. Classification methods require a multitemporal reference, which is difficult to extract from the raw data, therefore, these methods are not a practical solution \cite{demir2013classification}. Clustering techniques \cite{ghosh2011fuzzy,celik2010change,gong2011change,krylov2016false,liu2017change} are affected by parameter initialization, which may generate  local minima in the learning stage. In addition, the intrinsic brightness distortion in MS images yields inaccurate change maps \cite{song2018change}.

In order to reduce the effect of intra-class small variability and artifacts presented in MS images, we proposed a graph-based data fusion approach applied to CD. Our method extracts features from the dataset by finding eigenvectors of the normalized graph Laplacian and applying a mutual information based criteria to extract the relevant eigenvector that captures the change map. We validate our approach in two real cases: (i) a flooding, and (ii) a fire incident. Results show that our model reduces the effects of artifacts in the final change map, and it achieves low rates of false alarms in comparison to probabilistic threshold methods \cite{kittler1986minimum,zanetti2015rayleigh,zanetti2017theoretical}.

\vspace{-0.3cm}

\section{Graph based data fusion}
\label{sec:MMG}

\subsection{Graph}
A graph is a non linear structural representation of data, defined by $G = (V,E)$, where $G$ is the graph, $V$  is a set of nodes, and $E$ refers to the arcs or edges that explain the directed or undirected relationship between nodes. The edges have associated a weight $w_{i,j}$, that quantifies how strong the relationship is between nodes. The common measure used for each weight is a Gaussian kernel with standard deviation $\sigma$ \cite{alfredointroduccion}.

%\begin{equation}
%w_{i,j} = \exp\left(-\frac{d(V_{i},V_{j})^{2}}{\sigma^{2}}\right) \nonumber
%\end{equation}
%where, $d(V_{i},V_{j})$ is the distance between nodes and $\sigma$ is the standard deviation of all $d(V_{i},V_{j})$. A common application of graphs is the embedding of $G$ based on the Laplacian ($L$) matrix into space $\mathbb{R}^{m}$, keeping the graph nodes as close as they were on the input space. In short, the embedding of graph is given by the eigenproblem $L y = \lambda D y$ \cite{belkin2003laplacian}, where $L = D - W$, $W$ is known as the adjacency matrix or weights of the graph (each component is given by equation \eqref{eq:wij}), $D$ is a diagonal matrix which components are the degree of node ($di = \sum_{j} w_{i,j}$).\\

\subsection{Graph-based fusion for change detection (GBF-CD)}

\subsubsection{Nystr\"om extension}

Given the high number of pixels in an MS image, the computational cost of calculating the full matrix $\mathbf{W} \in \mathbb{R}^{N \times N}$ is extremely high (i.e an image with size $1280 \times 960$ is equivalent to $N = 1228800$). Therefore, an approximation of this matrix is computed through the Nystr\"om extension \cite{fowlkes2004spectral}: 

\begin{equation}\label{ec:nys}
\mathbf{W} = \kappa_{G}\left(
\begin{bmatrix}
\mathbf{d_{AA}} & \mathbf{d_{AB}}     \\
\mathbf{d_{AB}}^{\top} & \mathbf{C}     \\
\end{bmatrix}\right), \nonumber
\end{equation}

where $\kappa_{G}$ is a Gaussian kernel, $\mathbf{d_{AA}} \in \mathbb{R}^{n_{s} \times n_{s}}$, $\mathbf{d_{AB}} \in \mathbb{R}^{n_{s} \times (N - n_{s})}$ and $\mathbf{C} \in \mathbb{R}^{(N - n_{s}) \times (N - n_{s})}$. This method approximates $\mathbf{C}$ by choosing $n_{s}$ samples from the dataset of size $N$ ($n_{s} \ll N$), then $ \mathbf{W} \approx \mathbf{\widehat{W}} = \kappa_{G}\left(
\begin{bmatrix} \mathbf{d_{AA}} ; \mathbf{d_{AB}}\end{bmatrix}\right) ^{\top}$. Thus, the eigenvectors of the matrix $\mathbf{\widehat{W}}$, can be spanned by eigenvalues and eigenvectors of $\kappa_{G}(\mathbf{d_{AA}})$. Solving the diagonalization of $\kappa_{G}(\mathbf{d_{AA}})$  (eigenvalues $\lambda$ and eigenvectors $\mathbf{U}$: $\kappa_{G}(\mathbf{d_{AA}}) = \mathbf{U}^{\top} \mathbf{\Lambda} \mathbf{U}$), the eigenvectors of $\mathbf{\widehat{W}}$ can be spanned by $\mathbf{\hat{U}} = 
\begin{bmatrix}
\mathbf{U}  ; 
\kappa_{G}(\mathbf{d_{AB}})^{\top} \mathbf{U} \mathbf{\Lambda}^{-1}
\end{bmatrix} ^{\top}$. Since the approximated eigenvectors $\mathbf{\hat{U}}$ are not orthogonal, as explained in \cite{fowlkes2004spectral}, to obtain orthogonal eigenvectors we use $\mathbf{S} = \kappa_{G}(\mathbf{d_{AA}}) + \kappa_{G}(\mathbf{d_{AA}})^{- \frac{1}{2}}\kappa_{G}(\mathbf{d_{AB}})\kappa_{G}(\mathbf{d_{AB}})^{\top}\kappa_{G}(\mathbf{d_{AA}})^{- \frac{1}{2}}$.Then, by diagonalization of $\mathbf{S}$ ($\mathbf{S} = \mathbf{U_{s}} \mathbf{\Lambda_{s}} \mathbf{U_{s}}$) the final approximated eigenvectors of $W$ are given by:

\begin{equation}\label{ec:nys_Us}
\mathbf{\hat{U}} = 
\begin{bmatrix}
\kappa_{G}(\mathbf{d_{AA}})   \\
\kappa_{G}(\mathbf{d_{AB}})^{\top} \kappa_{G}(\mathbf{d_{AA}})^{- \frac{1}{2}} 	 \\
\end{bmatrix}\mathbf{U_{s}} \mathbf{\Lambda_{s}}^{- \frac{1}{2}}. \nonumber
\end{equation}

\subsubsection{Fusion step}

Based on the methodology introduced in\cite{iyer2017graph}, where a node is understood as a pixel (i.e. image from different bands or times, also a mix of both) and it is assumed that all of the modalities are co-registered, the fusion of multi-temporal data is carried out by the procedure described in \Cref{fig:MMG-chart}, in which for each instance of time ($X^{k}$) the Algorithm 1 is performed.

	\begin{figure*}[!h]
	\centering
	\includegraphics[scale=0.35]{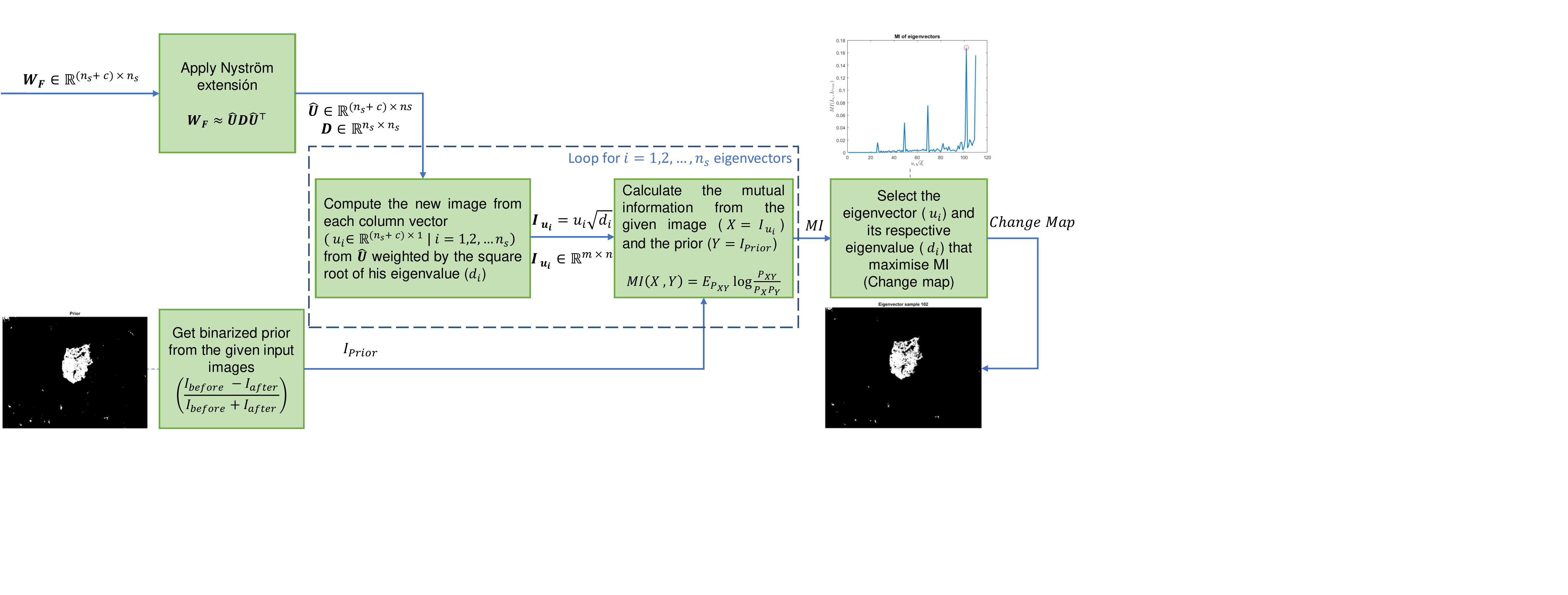}
	\caption{Change detection. Where, $\mathbf{\widehat{W}_{F}}$ is the fused graph, $\mathbf{\widehat{U}}$ is the approximate eigenvectors and $\mathbf{D}$ is the eigenvalues.}
	\label{fig:CD-chart}
\end{figure*}

\begin{algorithm}[!h]
	\SetAlgoLined
	\setstretch{0.9}
	\KwInput{Temporal images $\mathbf{X}^{k} \in \mathbb{R}^{m\times n}$, number of samples $n_s$}
	\KwOutput{Fused graph $\mathbf{W_{F}} \in \mathbb{R}^{(n_{s} + c)\times n_{s}}$}
	\textbf{Initialize:} $k = 1$, $N = m*n$\\
	
	\vspace{0.15cm}
	
	\While{$k \leq 2$}{
		
		\vspace{0.15cm}
		
		1) Take $n_{s}$ samples from $\mathbf{X}^{k}$ \\
		
		\quad $X_{AA}^{k} =$ sampler($\mathbf{X}^{k},n_s$), $\in \mathbb{R}^{n_{s}}$\\
		
		2) Find the complement $\overline{X}^{k} \in \mathbb{R}^{c}$ of $X_{AA}^{k}$ in $\mathbf{X}^{k}$.\\
		
		3) For each set $X_{AA}^{k}$ and $\overline{X}^{k}$ perform the pairwise distance between samples-samples ($\mathbf{d_{AA}}^{k} \in \mathbb{R}^{n_{s} \times n_{s}}$) and samples-complement ($\mathbf{d_{AB}}^{k} \in \mathbb{R}^{c \times n_{s}}$).\\
		
		\quad $\mathbf{d_{AA}}^{k} = \left\{\norm{x_{AA_{i}}^{k} - x_{AA_{j}}^{k}}_{2}\right\}_{i \hspace{0.2cm} j}^{n_{s} \space n_{s}}$, $\forall i \neq j$
		
		\quad $\mathbf{d_{AB}}^{k} = \left\{\norm{\overline{x}_{i}^{k} - x_{AA_{j}}^{k}}_{2}^{3}\right\}_{i \hspace{0.2cm} j}^{c \hspace{0.2cm} n_{s}}$, $\forall i \neq j$\\
		
		4)Apply the normalized graph laplacian ($\mathbf{\hat{D}}^{-\frac{1}{2}} \widehat{\mathbf{W}} \mathbf{\hat{D}}^{-\frac{1}{2}}$) on the distances by using the code in  \cite{fowlkes2004spectral} \\
		
		5) Apply a Gaussian kernel ($\kappa_{G}(.)$) on the normalized distances and build the approximated normalized laplacian matrix based on the Nystr\"om approximation.\\
		
		\quad$\mathbf{\widehat{W}}_{N}^{k} = \begin{bmatrix} \kappa_{G}(\mathbf{d_{AA}}^{k}) ; \kappa_{G}(\mathbf{d_{AB}}^{k})\end{bmatrix} ^{\top}$,  \\
		
		$k += 1$
	}
	
	$\mathbf{W_{F}} = min(\widehat{w}_{N_{i j}}^{k})$, with $i = 1,.., c ; j = 1,..,n_{s}$.
	
	\caption{GBF for temporal data}
\end{algorithm}
	
 \vspace{0.1cm}			
 In short, the algorithm output for one instance of time $X^{k}$ corresponds to the approximate normalized adjacency matrix ($\mathbf{\widehat{W}}_{N}^{k}$) \cite{fowlkes2004spectral}. Then, the fusion step consist of capturing the unique information given by each graph ($\mathbf{\widehat{W}}_{N}^{k}$) into one fused graph ($\mathbf{W_{F}}$). In order to achieve this fusion, we maximize the distance between pixels (i.e. choosing those pixels that preserve most of the information):

\begin{equation}
\mathbf{W_{F}} = min(\widehat{w}_{N_{ij}}^{k}), \text{with }k = [1 ,2], \nonumber
\label{eq:minW}
\end{equation}

where $w_{i,j}$ represents the weight of the node for each instance of time ($i = 1,2,\dots, c ; j = 1,2,\dots,n_{s}$). In this sense, the learning of this approach is data driven (uses a few $n_s$ samples to learn) and it will be restarted from scratch for each set of data.

\section{Application of GBF-CD for Change detection}

\subsection{Change detection scheme based on multi-temporal graph}

To obtain the change map from the multi-temporal graph (section \ref{sec:MMG}), we apply the scheme detailed in \Cref{fig:CD-chart}. Here, the purpose is to attain the best match that reflects the change produced by any source. To do this, we use the eigenvectors from the GBF-CD as descriptors of the change. Nevertheless, the number of eigenvectors is equal to the samples ($n_{s}$) taken from the instances of time. Hence, we estimate the mutual information to identify the relevant eigenvector that captures the global change.
The output in \Cref{fig:CD-chart} is a vector that contains the mutual information between the prior knowledge (difference image) and the change map generated by the eigenvectors of the GBF-CD. Finally the change map detected is the eigen-image ($I_{u_{i}}$)  that maximizes the mutual information.

% It is important to mention that an image given by an eigenvector of the GBF-CD ($I_{u_{i}}$) must be rearranged, because the samples are taken from different regions of the image. In the Nystr\"om extension the approximated vector comes from a matrix $\mathbf{\widehat{W}}$. However, the real locations of $\mathbf{d_{AA}}$ and $\mathbf{d_{AB}}$ correspond to the same location where the samples were taken from the image (samples and complement respectively). 

% Below is an example of how to insert images. Delete the ``\vspace'' line,
% uncomment the preceding line ``\centerline...'' and replace ``imageX.ps''
% with a suitable PostScript file name.
% -------------------------------------------------------------------------
%\begin{figure}[htb]
%
%\begin{minipage}[b]{1.0\linewidth}
%  \centering
%  \centerline{\includegraphics[width=8.5cm]{image1}}
%%  \vspace{2.0cm}
%  \centerline{(a) Result 1}\medskip
%\end{minipage}
%%
%\begin{minipage}[b]{.48\linewidth}
%  \centering
%  \centerline{\includegraphics[width=4.0cm]{image3}}
%%  \vspace{1.5cm}
%  \centerline{(b) Results 3}\medskip
%\end{minipage}
%\hfill
%\begin{minipage}[b]{0.48\linewidth}
%  \centering
%  \centerline{\includegraphics[width=4.0cm]{image4}}
%%  \vspace{1.5cm}
%  \centerline{(c) Result 4}\medskip
%\end{minipage}
%%
%\caption{Example of placing a figure with experimental results.}
%\label{fig:res}
%%
%\end{figure}

% To start a new column (but not a new page) and help balance the last-page
% column length use \vfill\pagebreak.
% -------------------------------------------------------------------------
%\vfill
%\pagebreak

\section{Experimental results and discussion}

\subsection{Databases}

\begin{figure}[h!]
	\centering
	\subfigure[Mulargia lake]{\includegraphics[width = 0.16 \textwidth]{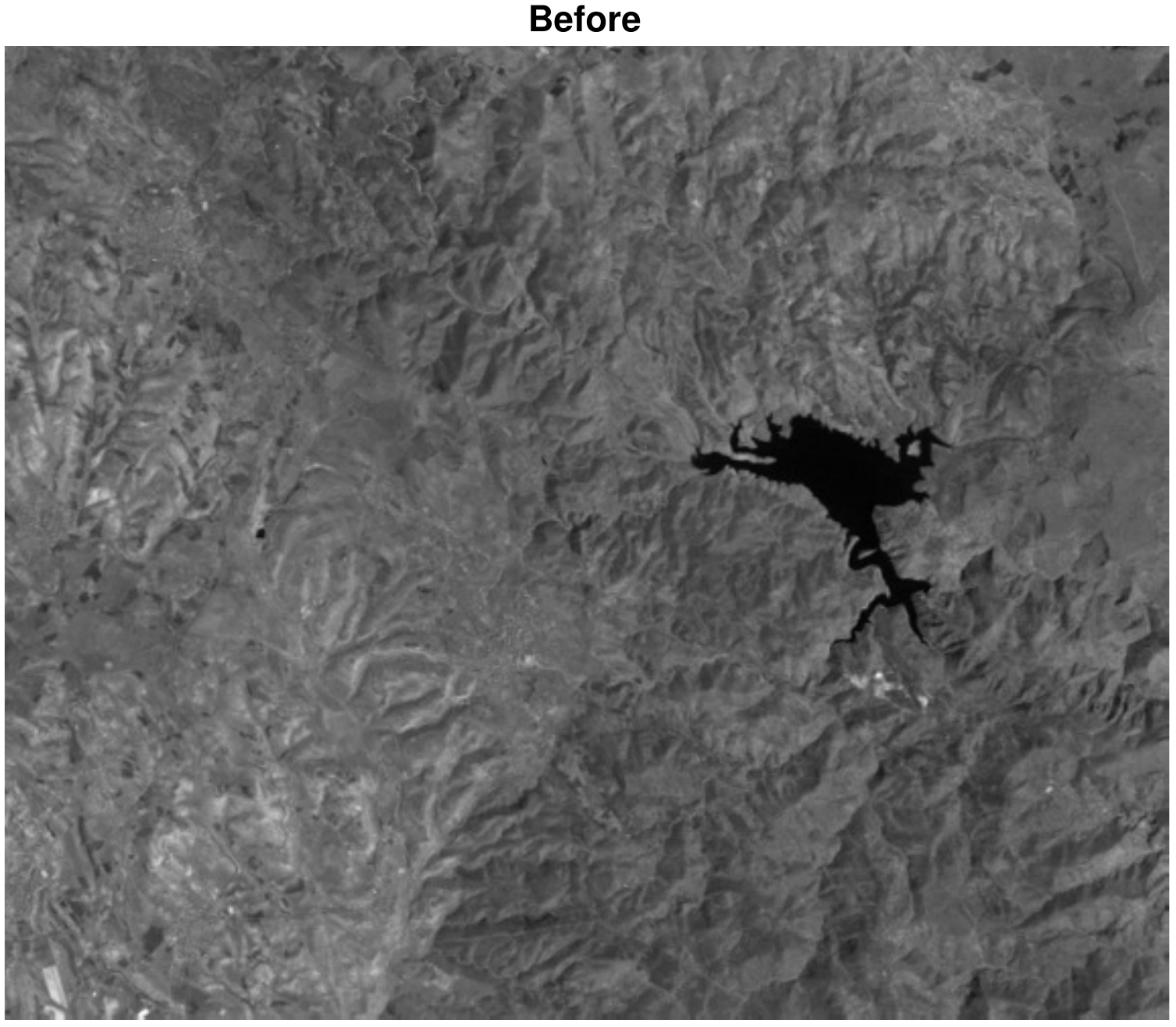}}
	\subfigure[Flooded Mulargia lake]{\includegraphics[width = 0.16 \textwidth]{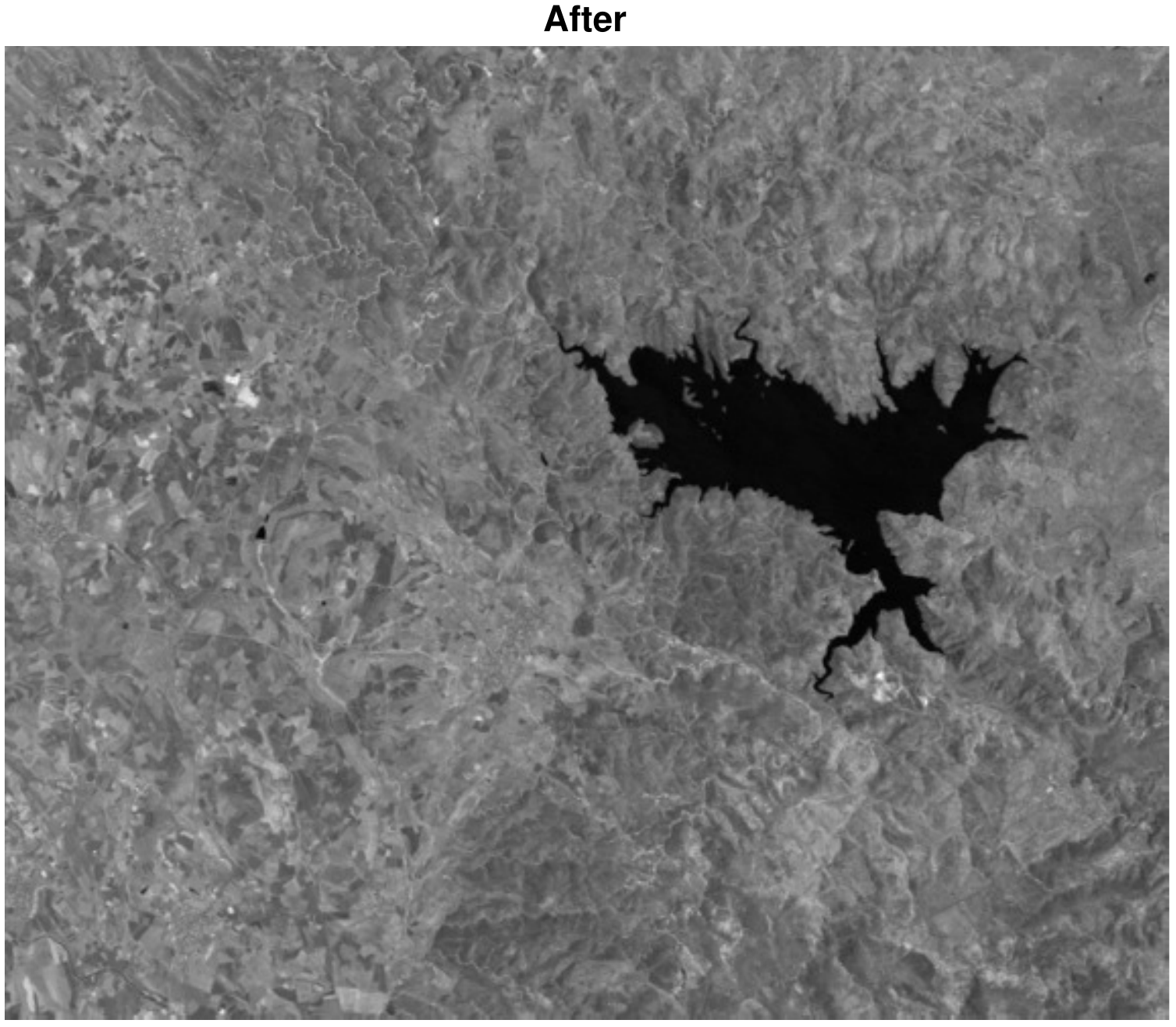}}\\
	\subfigure[Omodeo lake]{\includegraphics[width = 0.16 \textwidth]{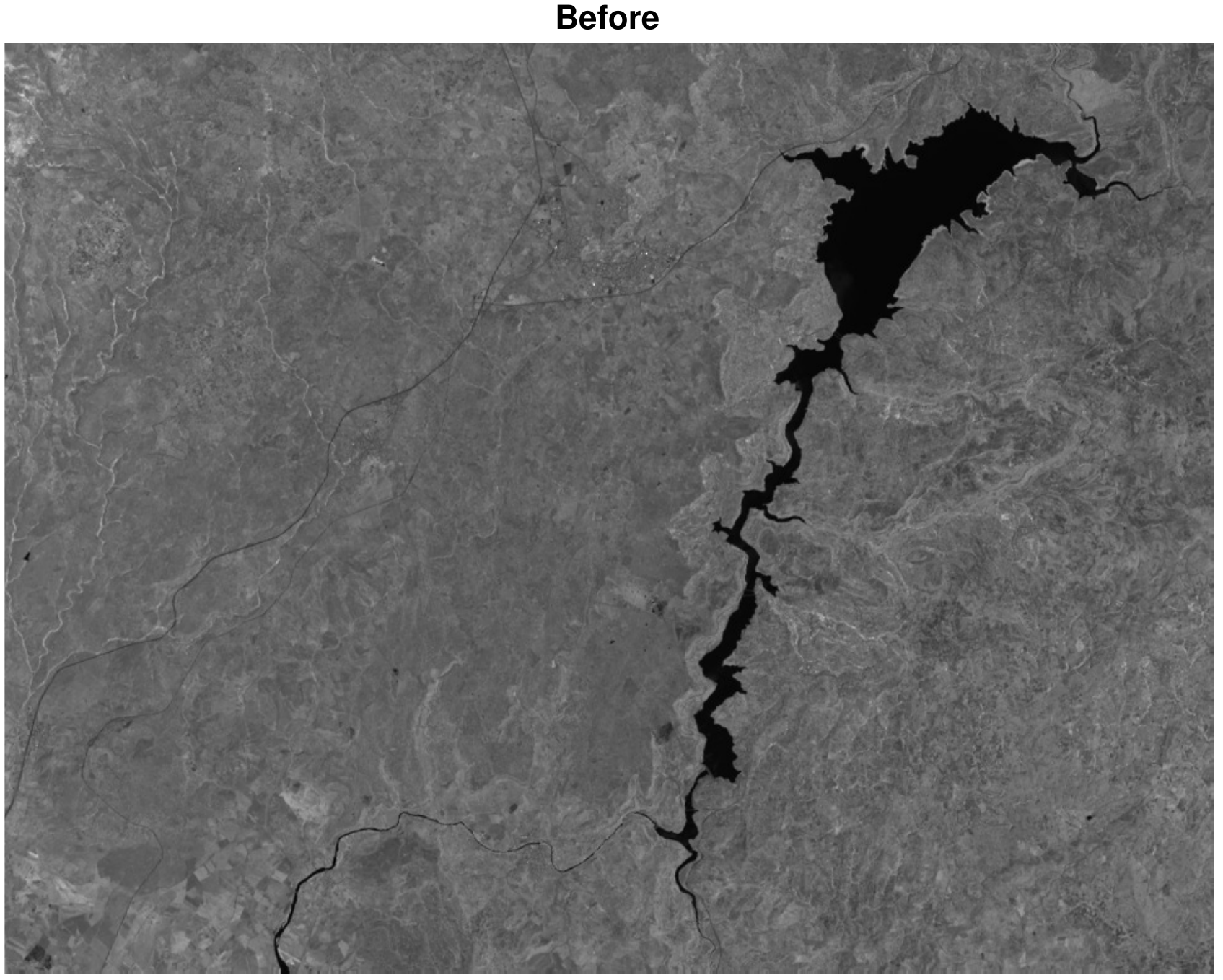}}
	\subfigure[Fire near Omodeo lake]{\includegraphics[width = 0.16 \textwidth]{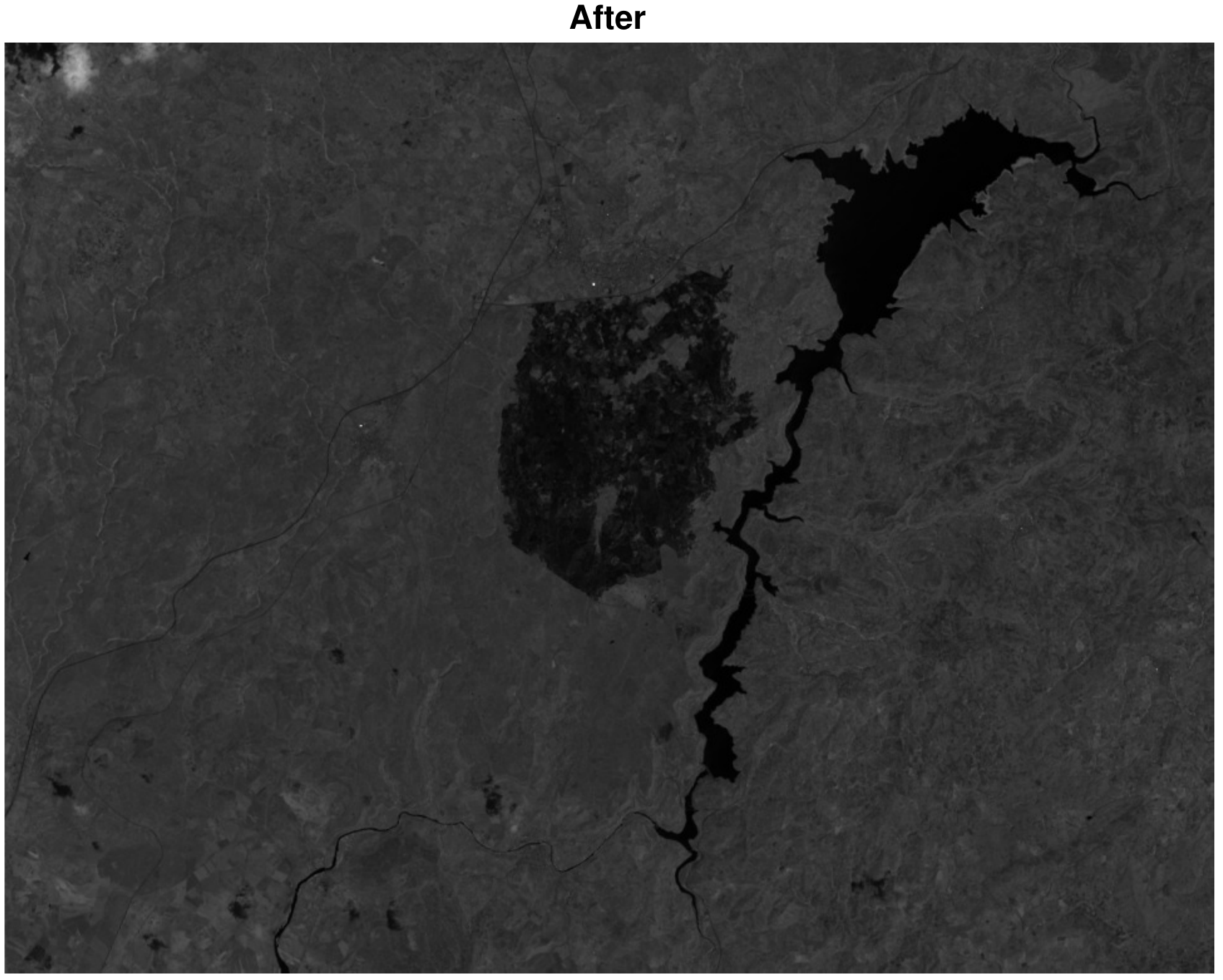}}
	\caption{Satellite images from the NIR band for the flood event (\emph{Dataset A}) and from the red band for fire event (\emph{Dataset B}).}
	\label{fig:dataset}
\end{figure}

Due to space limitations, we tested our approach on two datasets.: \textbf{Dataset A:} NIR band images (\Cref{fig:dataset} (a-b)) were acquired by the Thematic Mapper (TM) MS sensor of the Landsat-5 satellite. The scene represents an area including Lake Mulargia (Sardinia Island, Italy). The images consist of $573 \times 479$ pixels. The dates of acquisition were September 1995 (before the event) and July 1996 (after the event).
\textbf{Dataset B:} RED band images (\Cref{fig:dataset} (c-d)) were acquired by the Operational Land Image MS sensor of the Landsat-8 satellite. The area includes Lake Omodeo and a portion of Tirso River (Sardinia Island, Italy). The images consist of $965 \times 742$ pixels. The dates of acquisition were July 25, 2013 (before the event) and August 10, 2013 (after the event). 

\vspace{-0.4cm}

\subsection{Experimental set-up}

We compare the proposed GBF-CD with  state of the art methods: Rayleigh-Rice (rR) \cite{zanetti2015rayleigh}, Rayleigh-Rayleigh-Rice (rrR) \cite{zanetti2017theoretical}, and the classical Kittler–Illingsworth (KI) \cite{kittler1986minimum}. We evaluate relevant metrics in change detection such as: missed alarms (MA), false alarms (FA), precision (P), recall (R), Cohen's kappa (K) and overall error (OE).

The number of samples ($n_{s}$) was fixed at $92$ and the standard deviations ($\sigma$) for the kernels were $\sigma^{1}_{lake} = 2.5299 \times 10^{-10}$, $\sigma^{2}_{lake} = 1.5561 \times 10^{-10}$, $\sigma^{1}_{fire} = 2.793 \times 10^{-11}$ and $\sigma^{2}_{fire} = 1.6533 \times 10^{-10}$, where the superscripts ${1}$, ${2}$ stands for pre and post event respectively. We set these values through exhaustive grid-search using \emph{MatLab}$\textsuperscript{\textregistered} 2017a$. 

\begin{figure*}[h!]
	\centering
	\subfigure[KI]{\includegraphics[width = 0.20 \textwidth]{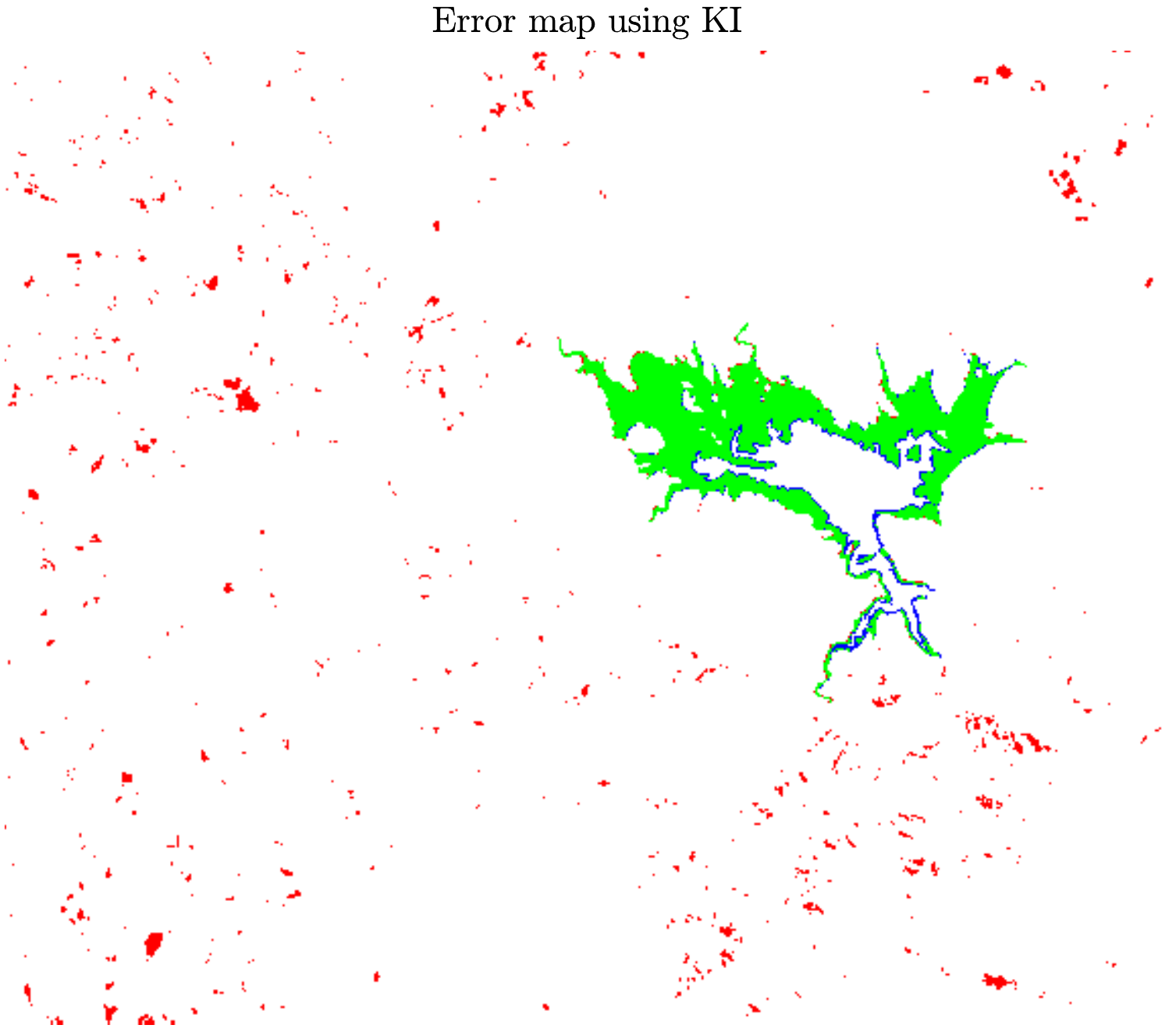}}
	\subfigure[rR]{\includegraphics[width = 0.20 \textwidth]{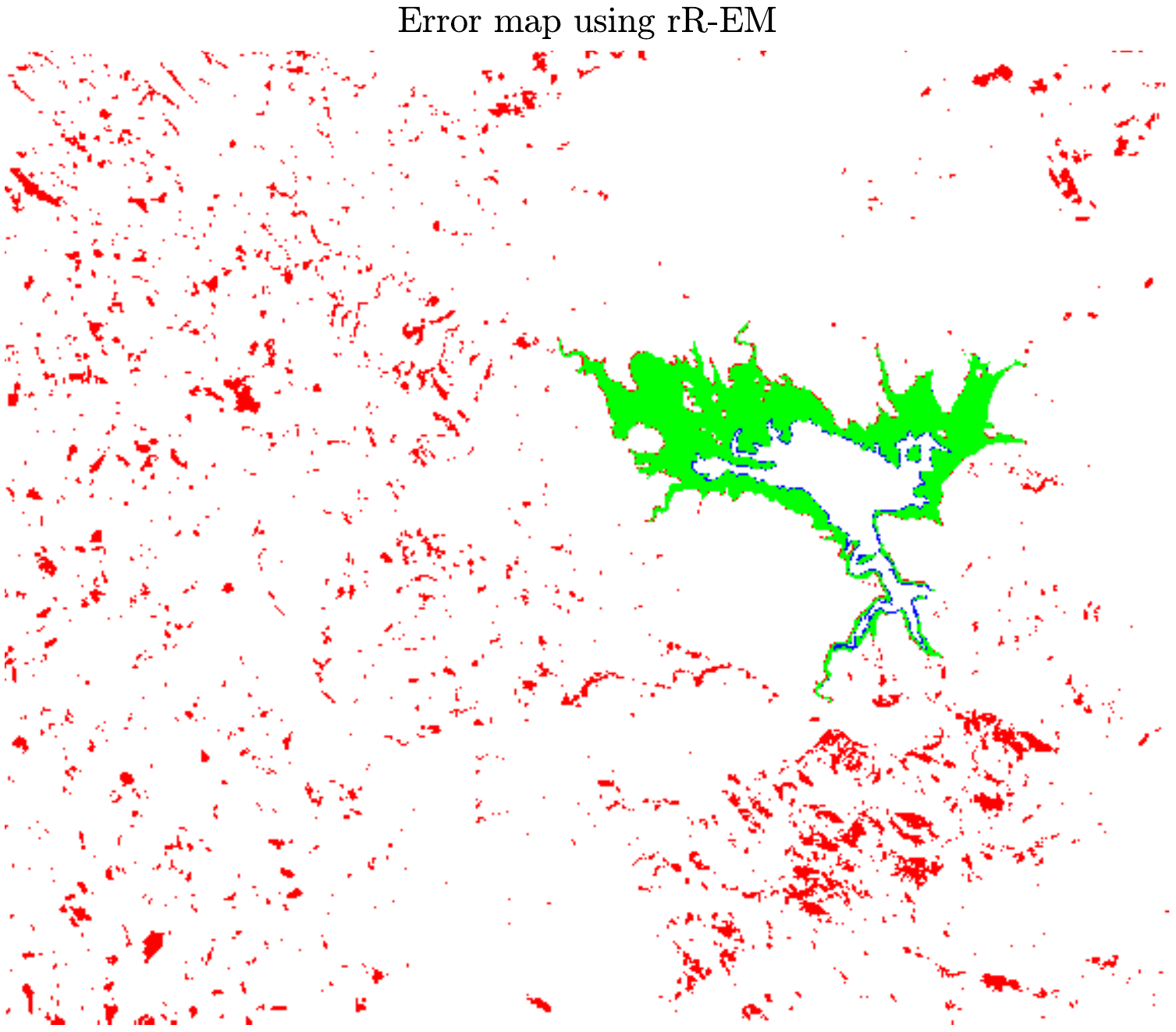}} 
	\subfigure[rrR]{\includegraphics[width = 0.20 \textwidth]{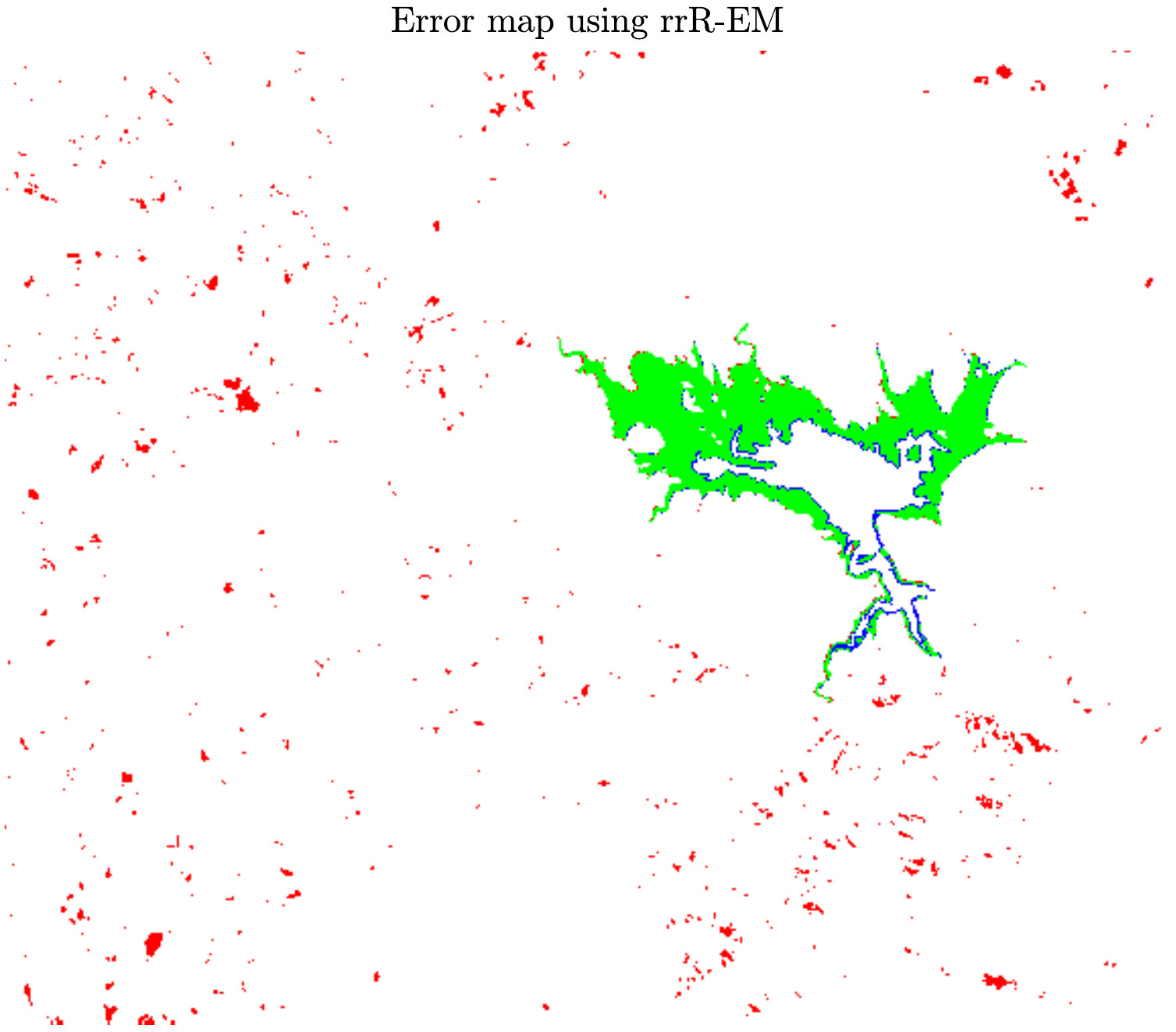}}
	\subfigure[GBF-CD]{\includegraphics[width = 0.20 \textwidth]{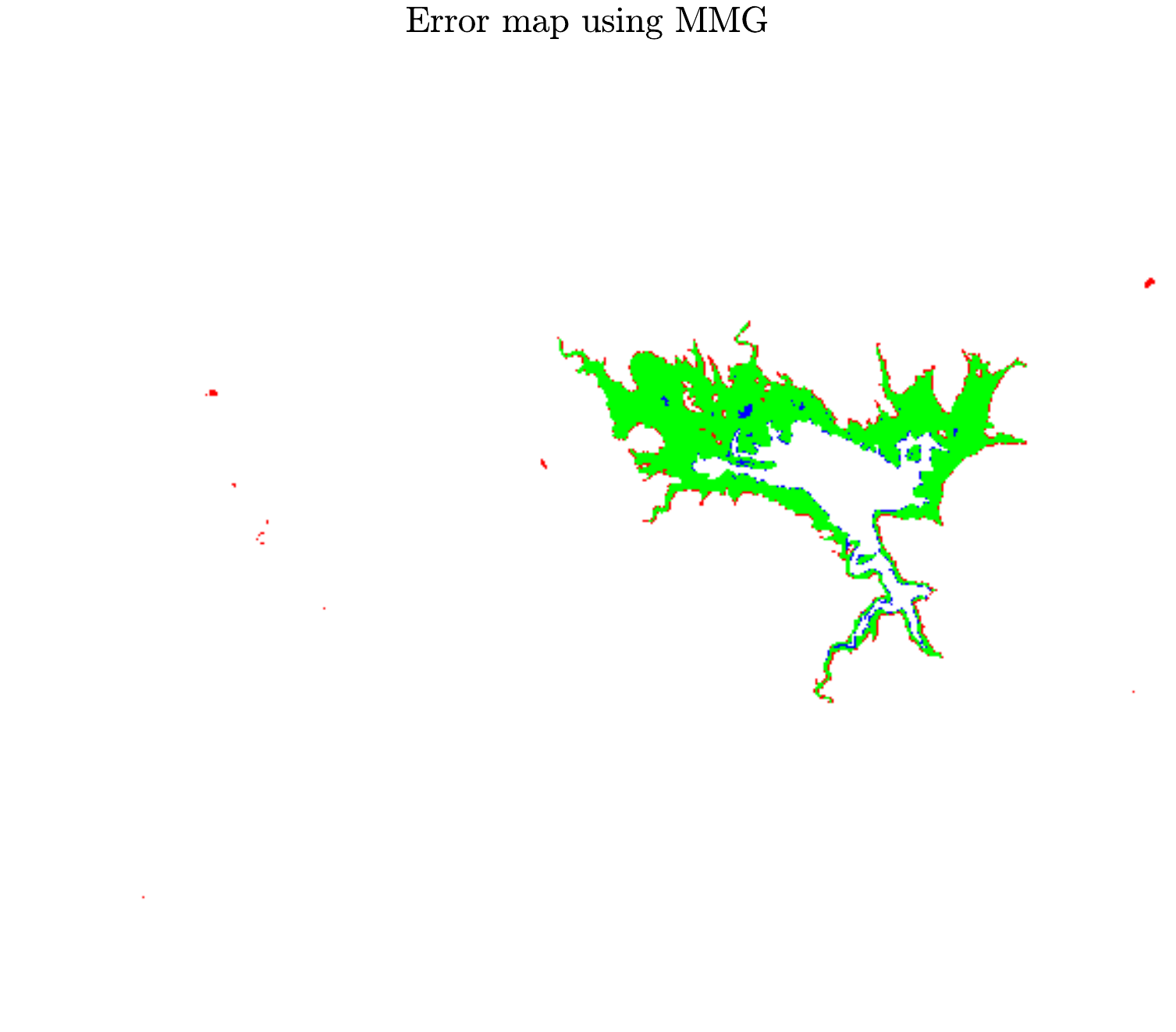}}\\ 
	\subfigure[KI]{\includegraphics[width = 0.20 \textwidth]{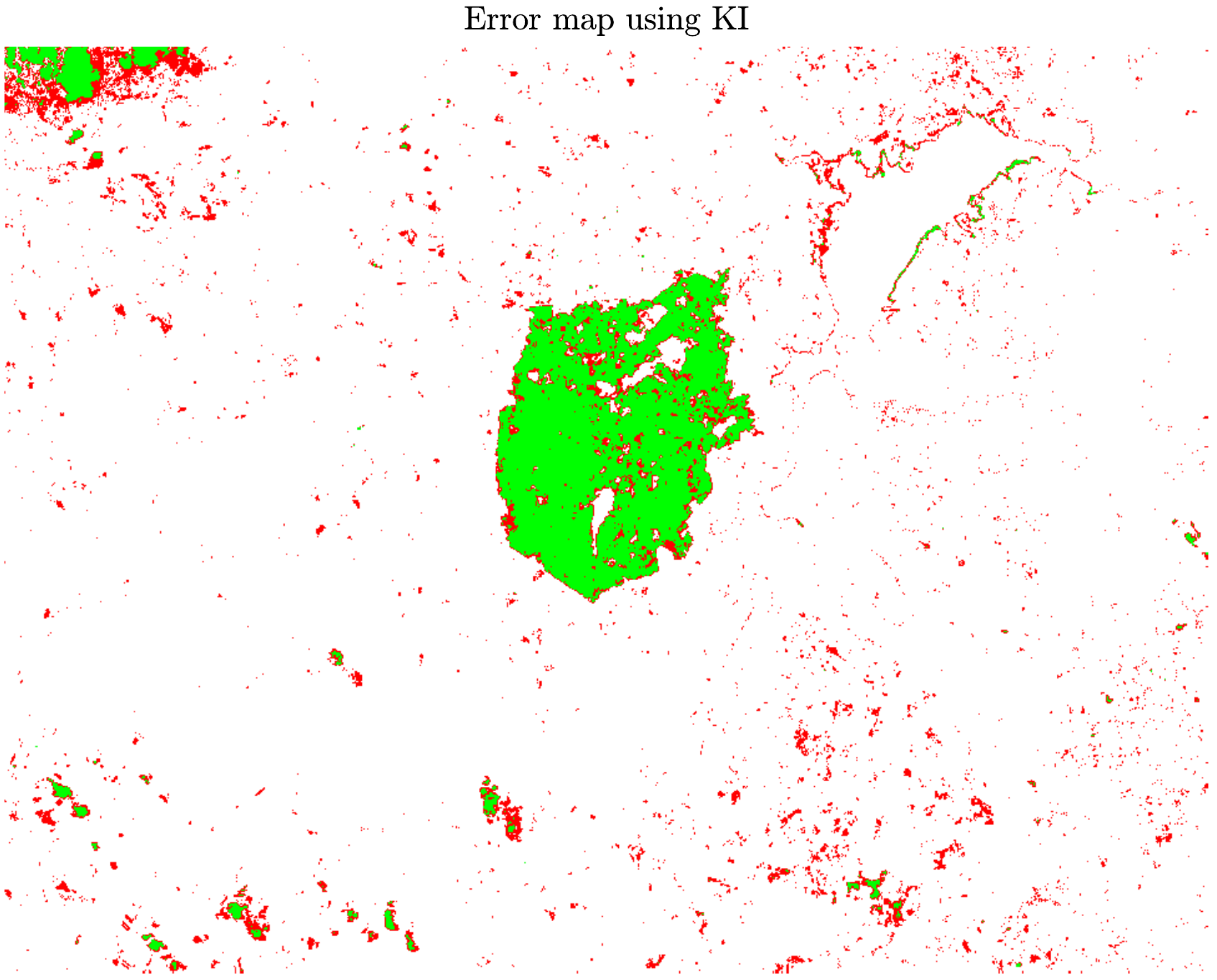}} 
	\subfigure[rR]{\includegraphics[width = 0.20 \textwidth]{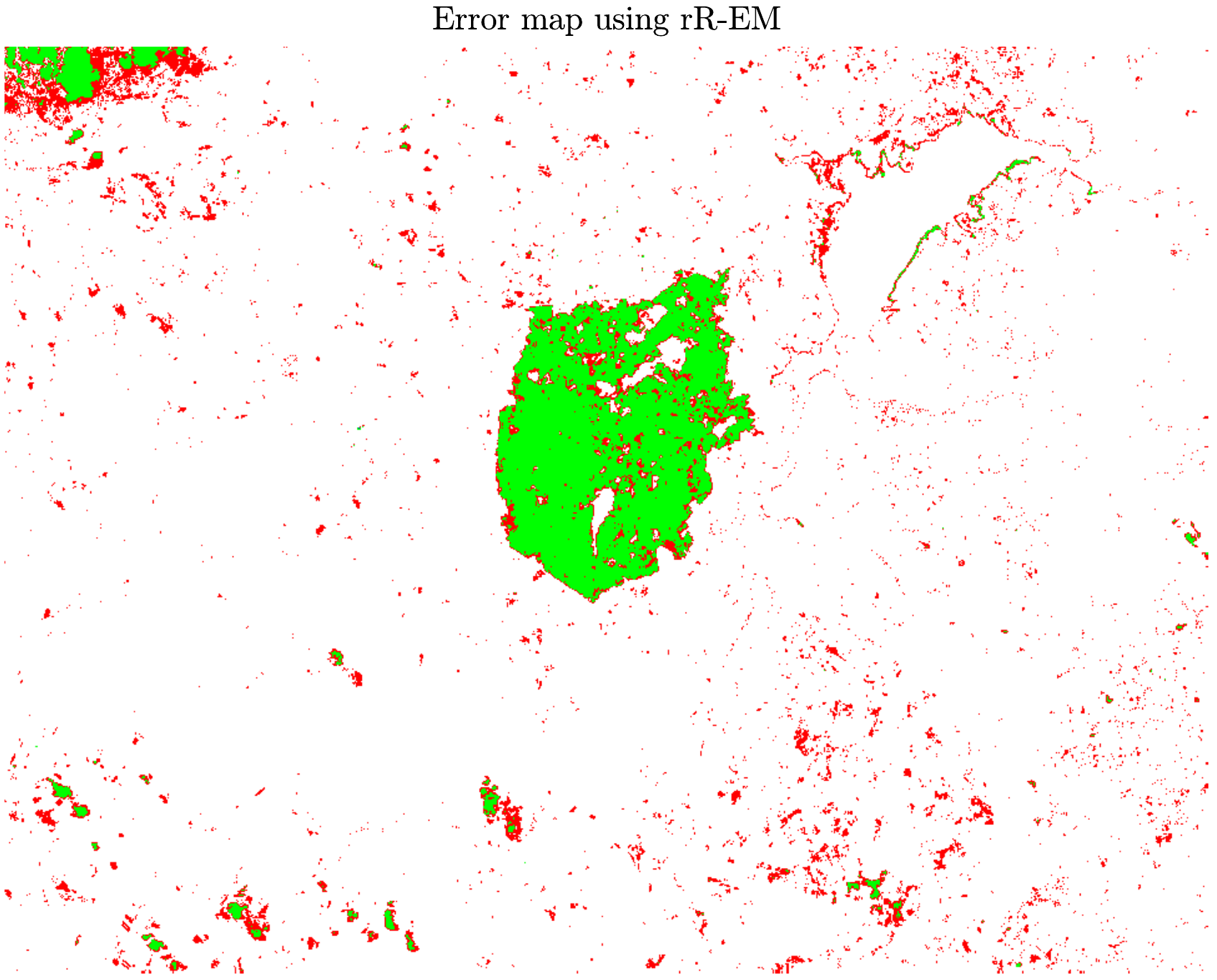}}
	\subfigure[rrR]{\includegraphics[width = 0.20 \textwidth]{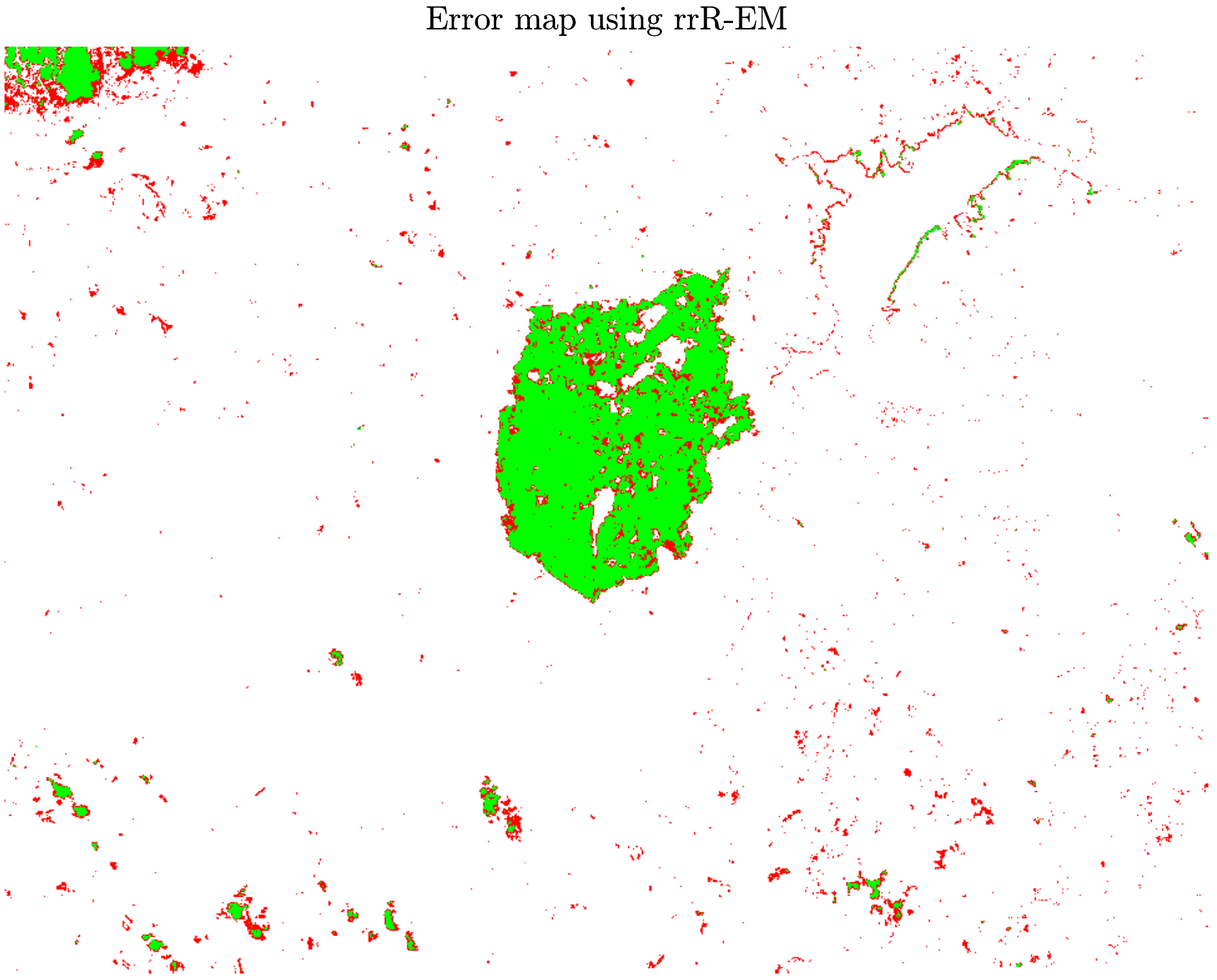}}
	\subfigure[GBF-CD]{\includegraphics[width = 0.20 \textwidth]{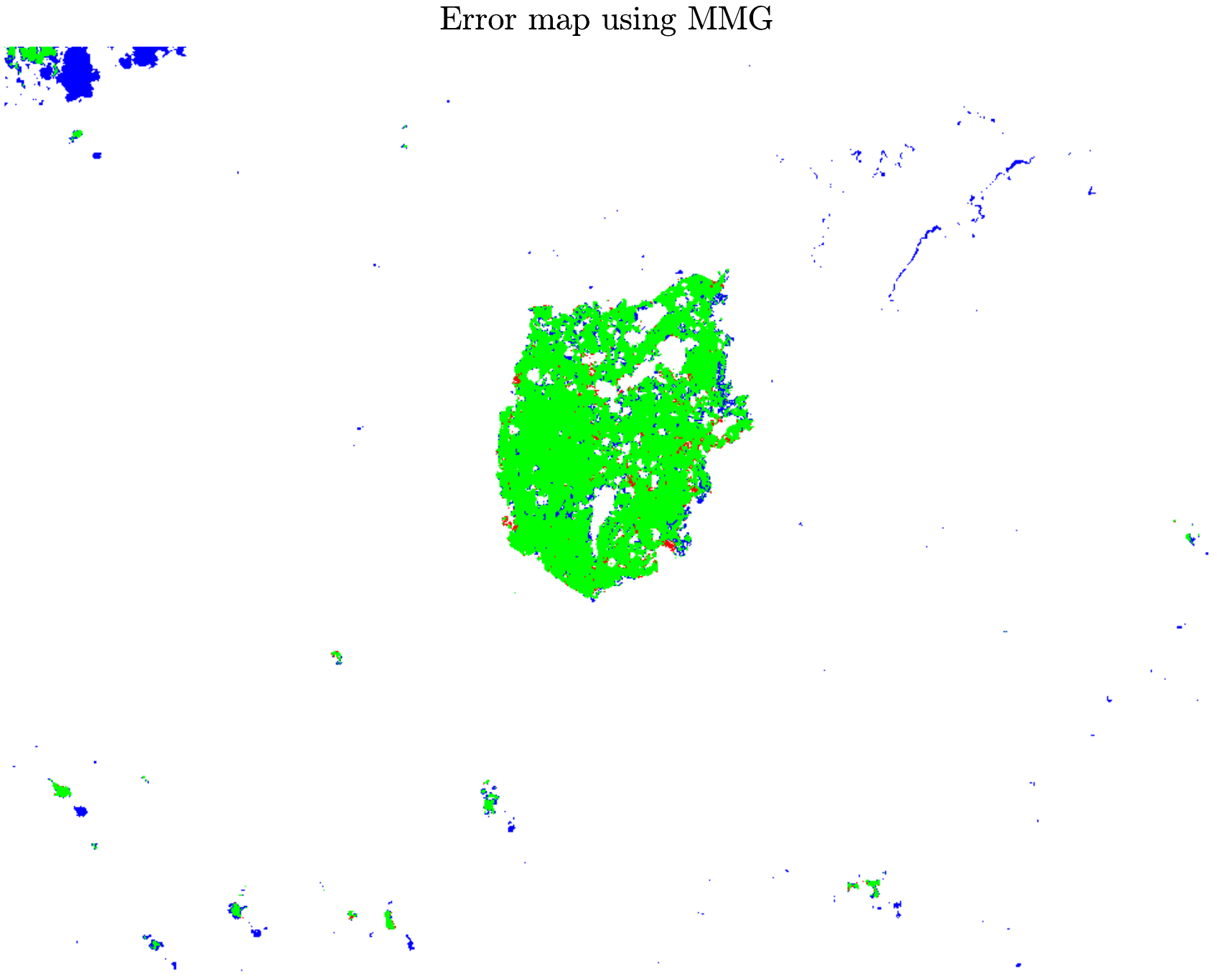}}
	\caption{Change map detected with respect to missed alarms (\textcolor{blue}{MA}), false alarms (\textcolor{red}{FA}) and corrrect changed pixels (\textcolor{green}{C}).}
	\label{fig:emaps}
\end{figure*}

\begin{table}[h!]
	\centering
	%	\footnotesize \onehalfspacing
	\caption{Model Performance for dataset A.}
	\resizebox{!}{0.7cm}
	{
	\begin{tabular}{@{}c|cccccc@{}}
		\hline
		\bfseries{Method} & \bf MA (\%) & \bf FA (\%) & \bf P & \bf R & \bf K & \bf OE (\%) \\ 
		\hline \hline
		KI \cite{kittler1986minimum}& 10.2425 & 1.0490 & 0.7229 & 0.8975 & 0.7941 & 1.3211 \\ \hline
		rR-EM \cite{zanetti2015rayleigh} & 5.7245 & 4.0147 & 0.4173 & 0.9427 & 0.5605 & 4.0653 \\ \hline
		rrR-EM \cite{zanetti2017theoretical} & 10.1440 & 1.0637 & 0.7203 & 0.8985 & 0.7928 & 1.3324 \\ \hline		
		\textbf{GBF-CD} & $\mathbf{4.8504}$ & $\mathbf{0.3120}$ & $\mathbf{0.9029}$ & $\mathbf{0.9515}$ & $\mathbf{0.9242}$ & $\mathbf{0.4463}$ \\ 
		\hline 
	\end{tabular}
	\label{tab:dataset_1}
	}
\end{table}

\begin{table}[h!]
	\centering
	%	\footnotesize \onehalfspacing
	\caption{Model Performance of dataset B.}
	\resizebox{!}{0.7cm}
	{
		\begin{tabular}{@{}c|cccccc@{}}
			\hline
			\bfseries{Method} & \bf MA (\%) & \bf FA(\%) & \bf P & \bf R & \bf K & \bf OE (\%) \\ 
			\hline \hline
			KI \cite{kittler1986minimum}& $\mathbf{0}$ & 3.4291 & 0.5903 & $\mathbf{1}$ & 0.7262 & 3.2676 \\ \hline
			rR-EM \cite{zanetti2015rayleigh} & 0.0029 & 3.7382 & 0.5693 & 0.9999 & 0.7080 & 3.5623 \\ \hline
			rrR-EM \cite{zanetti2017theoretical} & 0.0029 & 2.1449 & 0.6973 & 0.9999 & 0.8112 & 2.0440 \\ \hline			
			\textbf{GBF-CD} & 14.4217 & $\mathbf{0.1226}$ & $\mathbf{0.9718}$ & 0.8557 & $\mathbf{0.9059}$ & $\mathbf{0.7960}$\\ 
			\hline 
		\end{tabular}
		\label{tab:dataset_2}
	}
\end{table}

\Cref{tab:dataset_1} shows the results for dataset A. We observe our approach outperforms the comparison methods for all metrics. Similarly, \Cref{tab:dataset_2} tabulates the outcomes for dataset B. Although, the \textbf{KI} method achieves a perfect score for MA and recall (R), the \textbf{GBF-CD} outperforms the state-of-the-art methods in FA, P, K and OE.

Also, \Cref{fig:emaps} shows the behavior of each method in terms of MA (blue points), FA (red points) and correct changed pixels (green points). These results are remarkable, because we can see that probabilistic methods have a considerable number of FA in both datasets. Conversely, the \textbf{GBF-CD} deals well with this issue. FA is mostly generated by the nearly similar intensity of pixels between real changes regions and effects produced by the reflectance (i.e. weather variations, cloud density, daylight differences when the image was captured). The \textbf{GBF-CD} has some limitations. Firstly, for dataset A (see subfigure \ref{fig:emaps} (d)), we can observe that border of the change map is mainly composed by red points (FA). This is due to the neigboring pixels of the border have a similar intensity. Also, for dataset B (see figure \ref{fig:emaps} (h)), the \textbf{GBF-CD} is unable to detect minor changes in the edge of lake Omodeo and the artifact located in the upper-left corner. For this reason, there are some missed alarms (MA) represented by the blue points. 

Further work includes: (i) to decrease the dependence of the results with respect to the number of selected samples for the Nystr\"om extension, (ii) to select an alternative metric instead of Euclidean distance (ED) to increase the difference between intensities in MS images and avoid raising the ED to the power of three, (iii) to explore other kernel types.

%\section{RELATION TO PRIOR WORK}
%\label{sec:prior}
%
%The text of the paper should contain discussions on how the paper's
%contributions are related to prior work in the field. It is important
%to put new work in  context, to give credit to foundational work, and
%to provide details associated with the previous work that have appeared
%in the literature. This discussion may be a separate, numbered section
%or it may appear elsewhere in the body of the manuscript, but it must
%be present.
%
%You should differentiate what is new and how your work expands on
%or takes a different path from the prior studies. An example might
%read something to the effect: "The work presented here has focused
%on the formulation of the ABC algorithm, which takes advantage of
%non-uniform time-frequency domain analysis of data. The work by
%Smith and Cohen \cite{Lamp86} considers only fixed time-domain analysis and
%the work by Jones et al \cite{C2} takes a different approach based on
%fixed frequency partitioning. While the present study is related
%to recent approaches in time-frequency analysis [3-5], it capitalizes
%on a new feature space, which was not considered in these earlier
%studies."

\vspace{-0.6cm}

\blfootnote{To ensure the reproducibility of the proposed method, the code is publicly available at: \url{https://github.com/DavidJimenezS}}

\section{Conclusions}

In this paper, we introduced a change detection methodology (\textbf{GBF-CD}) based on graphs data fusion. Our main contribution is a ``data-driven'' framework. Our method models the dataset by finding eigenvectors of the normalized graph Laplacian and applying a mutual information based criteria to obtain the relevant eigenvector that captures the change map. Experimental results showed that \textbf{GBF-CD} outperformed probabilistic threshold methods when we evaluated several metrics (MA, FA, R, P, K, OE) in two real cases of change detection in remote sensing images. 

According to the previous results and analysis, we conclude that the \textbf{GBF-CD} is a promising and robust approach for detecting changes in remote sensing images.

\noindent \textbf{Acknowledgments}
This work was funded by the OMICAS program: ``Optimizaci\'on Multiescala In-silico de Cultivos Agr\'icolas Sostenibles (Infraestructura y validaci\'on en Arroz y Ca\~na de Az\'ucar)'', sponsored within the Colombian Scientific Ecosystem by The WORLD BANK, COLCIENCIAS, ICETEX, the Colombian Ministry of Education and the Colombian Ministry of Industry and Turism under GRANT ID: FP44842-217-2018.
% References should be produced using the bibtex program from suitable
% BiBTeX files (here: strings, refs, manuals). The IEEEbib.bst bibliography
% style file from IEEE produces unsorted bibliography list.
% -------------------------------------------------------------------------
\bibliographystyle{IEEEbib}
\bibliography{bib_df}

\end{document}